\documentclass[hyperfootnotes=false]{article}
\usepackage{graphicx,tabu,url}

\newcommand\blfootnote[1]{%
	\begingroup
	\renewcommand\thefootnote{}\footnote{#1}%
	\addtocounter{footnote}{-1}%
	\endgroup
}

\begin{document}

\title{Lens Distortion Rectification\\using Triangulation based Interpolation}

\author{Burak Benligiray \\ burakbenligiray@anadolu.edu.tr \and Cihan Topal \\ cihant@anadolu.edu.tr}

\date{}

\maketitle

\begin{abstract}
Nonlinear lens distortion rectification is a common first step in image processing applications where the assumption of a linear camera model is essential.
For rectifying the lens distortion, forward distortion model needs to be known.
However, many self-calibration methods estimate the inverse distortion model.
In the literature, the inverse of the estimated model is approximated for image rectification, which introduces additional error to the system.
We propose a novel distortion rectification method that uses the inverse distortion model directly.
The method starts by mapping the distorted pixels to the rectified image using the inverse distortion model.
The resulting set of points with subpixel locations are triangulated.
The pixel values of the rectified image are linearly interpolated based on this triangulation.
The method is applicable to all camera calibration methods that estimate the inverse distortion model and performs well across a large range of parameters.
\blfootnote{Source code: \url{https://github.com/bbenligiray/lens-distortion-triangulation}}
\end{abstract}

\section{Introduction}
\label{sec:intro}
Many computer vision and image processing methods assume a general pinhole camera model where linearity is preserved.
More specific applications, such as reconstruction or stereovision, benefit from precise estimation of camera parameters~\cite{hartley2003}.
Pinhole camera model is not directly applicable when the camera lens causes a nonlinear distortion in the image.
This distortion is commonly modeled using the Brown-Conrady polynomial model~\cite{brown1966} or Fitzgibbon's division model~\cite{fitzgibbon2001}.
These distortion models either define the mapping from distorted coordinates to rectified coordinates, or vice versa.
Parameters for either of these versions of the methods are estimated with distortion calibration methods.
However, a problem arises if the application requires the inverse mapping.
Since the mapping function is not analytically invertible, the inverse is required to be approximated~\cite{mallon2004}.

The conventional method of distortion rectification of an image requires the mapping from rectified coordinates to distorted coordinates.
We will name the parameters that define this mapping as \textit{(forward) distortion parameters}, as this is the most common objective of distortion calibration.
The parameters that define the mapping from distorted coordinates to rectified coordinates will be referred to as \textit{inverse distortion parameters}.
In calibration cases where a set of distorted and rectified point correspondences can be generated using the estimated intrinsic and extrinsic camera parameters, parameters of the distortion model can be estimated in both ways.
However, in self-calibration of lens distortion, the options are limited.
The common method is to rectify detected features with different inverse distortion parameters iteratively, until the resulting rectified features are believed to belong to a correctly rectified image~\cite{devernay2001,brauer2001,wang2009}.
The general assumption of line straightness methods is that the correctly rectified image contains more linear features compared to sub-optimally rectified images.
The method of quantifying the linearity of the features in an image is one of the main differences among line straightness methods.
When applying such methods, the estimated inverse distortion parameters are not used for rectifying the image directly.
Instead, the distortion parameters are needed to be approximated using the estimated inverse parameters.
Similar to this problem, points from the distorted image cannot be mapped to the rectified image knowing only the distortion parameters.
This mapping is defined by the inverse distortion model.

The distortion parameters can be directly inverted if only a single distortion parameter is used~\cite{bukhari2013,cucchiara2003,devernay2001}.
If multiple parameters are used, the inverse of the known distortion function is approximated.
Doing so will introduce additional error to the system.
We propose a novel method for rectifying the lens distortion using the inverse parameters directly, thus skipping the approximation step.
The inverse distortion model maps pixels of the distorted image to subpixel locations on the rectified image.
Since not all pixels of the rectified image are covered by this mapping, blank regions appear on the rectified image.
In the proposed method, the rectified point set is triangulated, such that a pixel inside this point set's convex hull will be surrounded by a triangle whose vertices are three mapped points.
These three mapped points are used to linearly interpolate the value at the pixel.
By doing this for every pixel, the image can be rectified with no blank regions.

Other methods approximate the solution with some error~\cite{gonzalez2011,alvarez2009,mallon2004,heikkila2000,heikkila1997,wei1994} and interpolate to rectify the distortion, meanwhile our method does the inversion implicitly in a single interpolation operation.
Furthermore, our method can work with more complex approximations of lens distortion without needing any modification or additional processing time.

\section{Lens Distortion Model}
\label{sec:distortion}

For self-calibration of lens distortion, polynomial~\cite{gonzalez2011,alvarez2009,grammatikopoulos2007,ahmed2005,mallon2004,cucchiara2003,thormahlen2003,devernay2001,brauer2001,prescott1997} and division~\cite{bukhari2013,wang2009,strand2005,brauer2002,fitzgibbon2001} radial lens distortion models are used.
Division model is generally approximated with a single parameter, exception being~\cite{brauer2002}.
Methods that use polynomial model use either one~\cite{ahmed2005,cucchiara2003,devernay2001,brauer2001} or two parameters~\cite{gonzalez2011,alvarez2009,grammatikopoulos2007,mallon2004,thormahlen2003,prescott1997}.
Out of these methods, only Ahmed and Farag's estimates a parameter of tangential lens distortion \cite{ahmed2005}.
Many methods estimate a distortion center different from the principal point~\cite{bukhari2013,wang2009,ahmed2005,brauer2002,devernay2001,prescott1997}, which can be said to model some aspects of tangential distortion, along with the ratio of vertical and horizontal focal lengths~\cite{devernay2001}.
We will use the polynomial distortion model with only two radial distortion parameters for brevity.
The proposed method for distortion rectification can be easily extended to division model, tangential distortion modeling and additional distortion parameters.

Let us assume $(\tilde{x}_{d},\tilde{y}_{d})$ are distorted screen coordinates and $(\tilde{x}_{u},\tilde{y}_{u})$ are rectified screen coordinates.
These coordinates can be normalized by subtracting the distortion center, $(c_{x},c_{y})$, thus yielding $(x_{d},y_{d})$ and $(x_{u},y_{u})$, respectively.
The distances of the normalized points from distortion centers give the radii, $r_{d}$ and $r_{u}$.
According to this notation, the radial distortion can be modeled in two ways:

\begin{equation}
\label{eq:forwarddistortion}
r_{d}=r_{u}(1+\kappa_{1}r_{u}^{2}+\kappa_{2}r_{u}^{4})
\end{equation}
\begin{equation}
\label{eq:inversedistortion}
r_{u}=r_{d}(1+\kappa_{1}^{\prime}r_{d}^{2}+\kappa_{2}^{\prime}r_{d}^{4})
\end{equation}

Eq.~\ref{eq:forwarddistortion} will be referred to as the distortion model, as it models the mapping from non-distorted points to distorted points.
Eq.~\ref{eq:inversedistortion} will be referred to as the inverse distortion model, as it does the opposite.
The effect of these distortion models on $x$ and $y$ coordinates are as follows:

\begin{equation}
\label{eq:forwarddistortionXY}
x_{d}=x_{u}(1+\kappa_{1}r_{u}^{2}+\kappa_{2}r_{u}^{4})  
y_{d}=y_{u}(1+\kappa_{1}r_{u}^{2}+\kappa_{2}r_{u}^{4})
\end{equation}

\begin{equation}
\label{eq:inversedistortionXY}
x_{u}=x_{d}(1+\kappa_{1}^{\prime}r_{d}^{2}+\kappa_{2}^{\prime}r_{d}^{4})  
y_{u}=y_{d}(1+\kappa_{1}^{\prime}r_{d}^{2}+\kappa_{2}^{\prime}r_{d}^{4})
\end{equation}

$\kappa_{1}$ and $\kappa_{2}$ in Eq.~\ref{eq:forwarddistortion} are the distortion parameters and $\kappa_{1}^{\prime}$ and $\kappa_{2}^{\prime}$ in Eq.~\ref{eq:inversedistortion} are the inverse distortion parameters.
In a majority of applications, this distinction is not made and the more convenient model is used directly.
When using a line straightness based method, only the distorted coordinates of the detected features are known.
Generating rectified coordinates using a set of parameters uses the inverse distortion model.
Thus, inverse distortion parameters are found when such methods are used.

\section{Distortion Rectification}
\label{sec:rectification}

Conventional distortion rectification uses the forward distortion model, which is the model that generates distorted coordinates, using the distortion parameters and rectified coordinates (Eq.~\ref{eq:forwarddistortionXY}).
Assume we have an empty image that will store the result of the rectification.
Obviously, each pixel of this image needs to be filled.
Distortion parameters ($\kappa_{i}$) and rectified coordinates ($x_{u},y_{u}$) are known, thus the distorted coordinates ($x_{d},y_{d}$) can be found.
The empty pixel in the rectified image will correspond to a subpixel location in the distorted image.
The value of the pixel is commonly estimated with bilinear interpolation, which will use the four surrounding pixels' values.
Doing this operation for each pixel in the empty image will result in a complete rectified image.
Now let us discuss the cases in which only the inverse distortion parameters are known.

\subsection{Inversion by Newton-Raphson Method}

Newton-Raphson method is used to find roots of a function by iteration.
Using Eq.~\ref{eq:inversedistortion}:

\begin{equation}
\label{eq:nm1}
f(r_{d})=r_{d}+\kappa_{1}^{\prime}r_{d}^{3}+\kappa_{2}^{\prime}r_{d}^{5}-r_{u}
\end{equation}
\begin{equation}
\label{eq:nm2}
f^{\prime}(r_{d})=1+3\kappa_{1}^{\prime}r_{d}^{2}+5\kappa_{2}^{\prime}r_{d}^{4}
\end{equation}

This function's derivative exists for possible roots (Eq.~\ref{eq:nm2}), thus the method is applicable.
The method approximates to the solution by iteration in the following manner:

\begin{equation}
\label{eq:nm3}
{r_{d}}_{n+1}={r_{d}}_{n}-\frac{f({r_{d}}_{n})}{f^{\prime}({r_{d}}_{n})}
\end{equation}

A good starting point for $r_{d}$ can be $r_{u}$, as they will be close when the distortion is weak, and there is no better choice for cases in which the distortion is strong.
After estimating $r_{d}$, we can find ($x_{d},y_{d}$) using Eq.~\ref{eq:inversedistortionXY}.
By mapping pixel coordinates of rectified image to subpixel locations in the distortion image using this method, conventional distortion rectification can be done~\cite{gonzalez2011}.
The downside of this method is that this iteration must be done for each pixel and convergence will take longer for stronger distortions.

\subsection{Taylor Expansion Based Inverse Model}

Taylor expansion~\cite{heikkila2000} and implicit rational polynomials~\cite{alvarez2009,heikkila1997,wei1994} are utilized in some studies, yet they are argued to be unstable by Mallon and Whelan.
Instead, Taylor expansion method is refined to be~\cite{mallon2004}:

\begin{equation}
\label{eq:mallondistortion}
p_{u}=p_{d}-p_{d}\left(\frac{\alpha_{1}r_{d}^{2}+\alpha_{2}r_{d}^{4}+\alpha_{3}r_{d}^{6}+\alpha_{4}r_{d}^{8}}{1+4\alpha_{5}r_{d}^{2}+6\alpha_{6}r_{d}^{4}}\right)
\end{equation}

In this form, this model estimates the inverse model, knowing the forward model.
However, it can also be used to estimate the forward model, knowing the inverse distortion model, which is the problem we have defined.

\section{Proposed Method: Rectification using Inverse Distortion Model}

\begin{figure}
	\centering
	\includegraphics[width=0.5\linewidth]{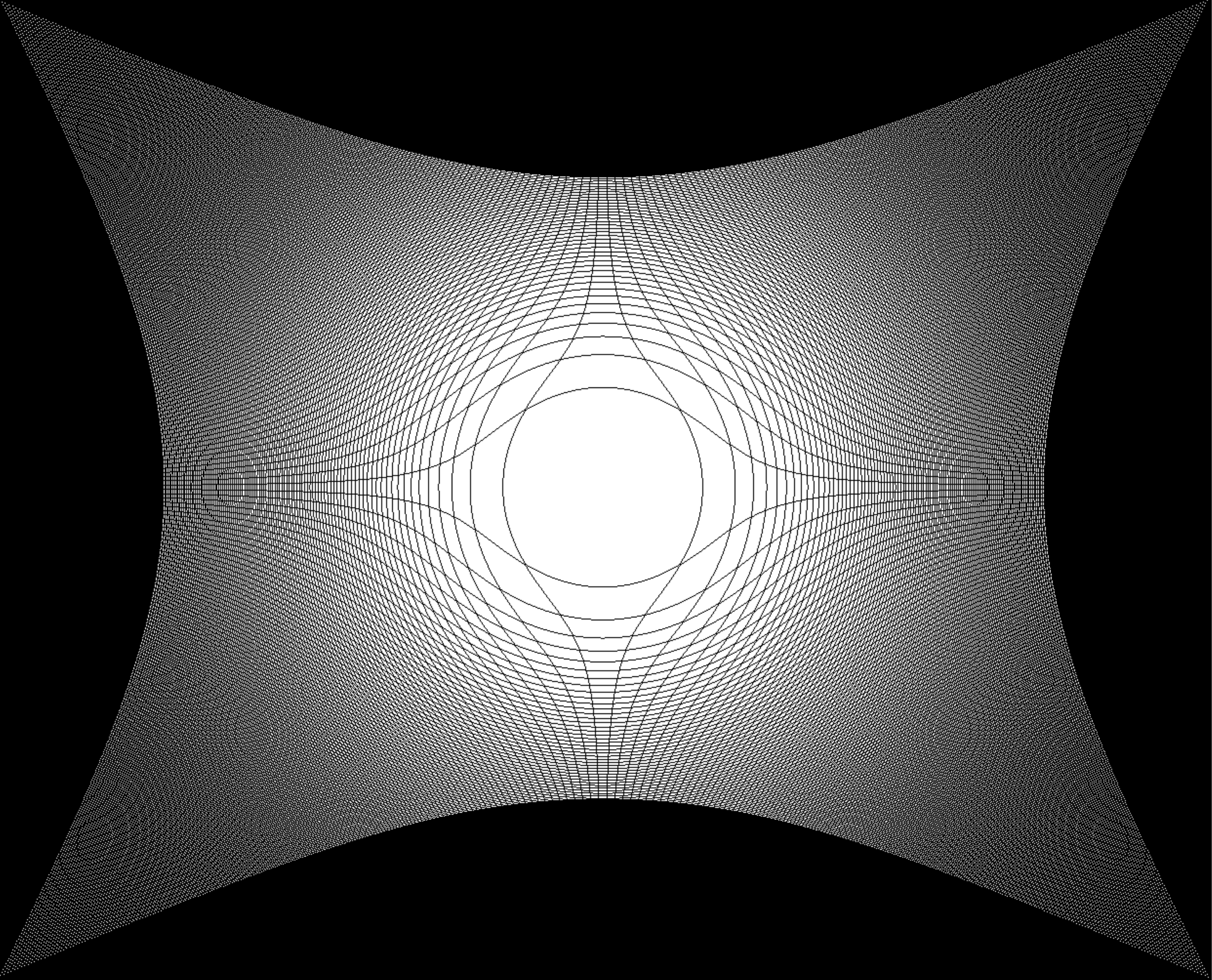}
	\caption{Rectification of a distorted fully white image using the inverse distortion function with nearest neighbor interpolation.}
	\label{fig:blank}
\end{figure}

Previously mentioned methods aim to approximate the distortion model from the estimated inverse distortion model, which will introduce additional error to the system, as will be shown in our experimental results.
Instead, directly using the inverse distortion model would have been preferable.
To do so, an approach similar to conventional distortion rectification can be followed.
Let us start with an empty rectified image.
For each pixel in the distorted image, the corresponding pixel in the empty image can be filled using the inverse distortion model.
The problem with this approach is that not all pixels in the rectified image are guaranteed to be filled, as addressed in other studies~\cite{gonzalez2011,cucchiara2003}.
Certain patterns of voids will appear at the rectified image (see Figure~\ref{fig:blank}).
Prescott and McLean vote on the surrounding four pixels of the mapped coordinates instead of using nearest neighbor, yet doing so still does not guarantee the elimination of all voids in the image~\cite{prescott1997}.

The problem can simply be defined as interpolating a nonlinearly scattered sparse image.
This is not a very common issue in image processing, but the solution can be derived from a similar problem.
Image sequence super-resolution aims to create a high resolution image from multiple low resolution images.
After fusing the information from lower resolution images, the result is a pseudo-random scattering of points that do not fit the grid of the output image.
Lertrattanapanich and Bose propose using Delaunay triangulation for solving this 2D interpolation problem~\cite{lertrattanapanich2002}.
This approach is applicable to our case.

\begin{figure}
	\centering
	\includegraphics[width=0.4\linewidth]{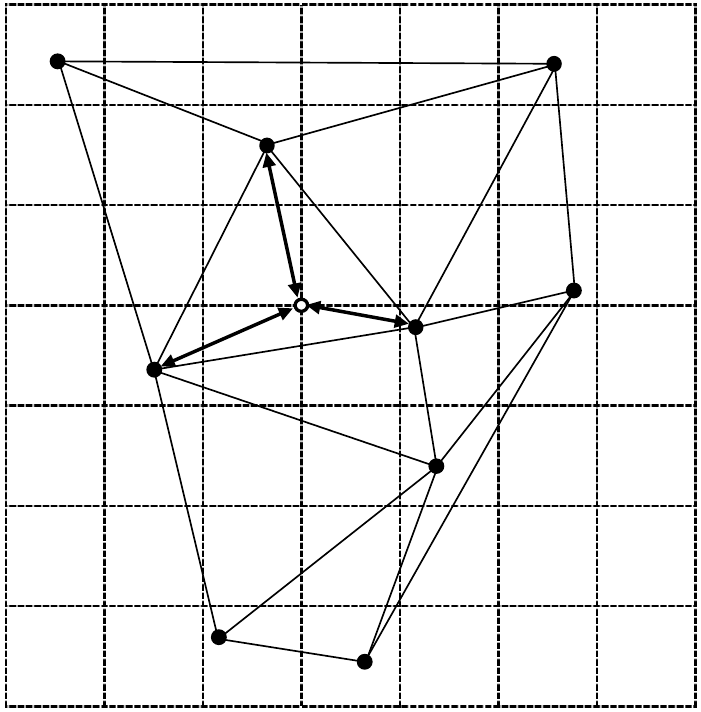}
	\caption{Triangulation based 2D interpolation.
		Corners of the grid represent the pixels of the rectified image.
		Filled circles are distorted pixels mapped to the rectified image.
		The empty circle represents a single pixel in the rectified image.
		The vertices of the surrounding triangle will vote on the value of this pixel, weighted by their Euclidean distances.}
	\label{fig:grid}
\end{figure}

The pixels from the distorted image are mapped to the rectified image plane using Eq.~\ref{eq:inversedistortionXY}, which will produce a set of points in subpixel locations.
Delaunay triangulation~\cite{lee1980} is applied to this set of points.
Each pixel of the rectified image will be surrounded by a triangle.
Using the values of the surrounding triangle's vertices, the value at the pixel can be estimated.
This estimation can be done linearly as proposed by Dyn et al.~\cite{dyn1990}, which resembles Phong's illumination model~\cite{phong1975}.
In this case, each pixel will be voted on by three points, instead of four as is the case in bilinear interpolation done in conventional distortion rectification (see Figure~\ref{fig:grid}).

The effect of using a different interpolation method is rather unpredictable.
The used triangulation method does not guarantee finding the closest three points to the pixel whose value will be interpolated.
Especially where points are sparser, suboptimal interpolations can be expected.
In such cases, any type of linear interpolation will fail to estimate the values correctly.
On the other hand, not approximating the inverse model is a certain advantage.

\section{Experimental Results}
\label{sec:experiment}

The proposed method directly uses the inverse distortion parameters to rectify the lens distortion.
Other methods approximate the distortion parameters using the inverse parameters and apply conventional distortion rectification.
Since our method's output is a rectified image, doing experiments with randomly generated point data and measuring the geometric error is not possible.
Instead, we will apply artificial lens distortion to an image.
Then, the distortion will be rectified using alternative methods and differences from the original image will be evaluated.

\begin{figure}
	\begin{minipage}[b]{0.48\linewidth}
		\centering
		\centerline{\includegraphics[width=6cm]{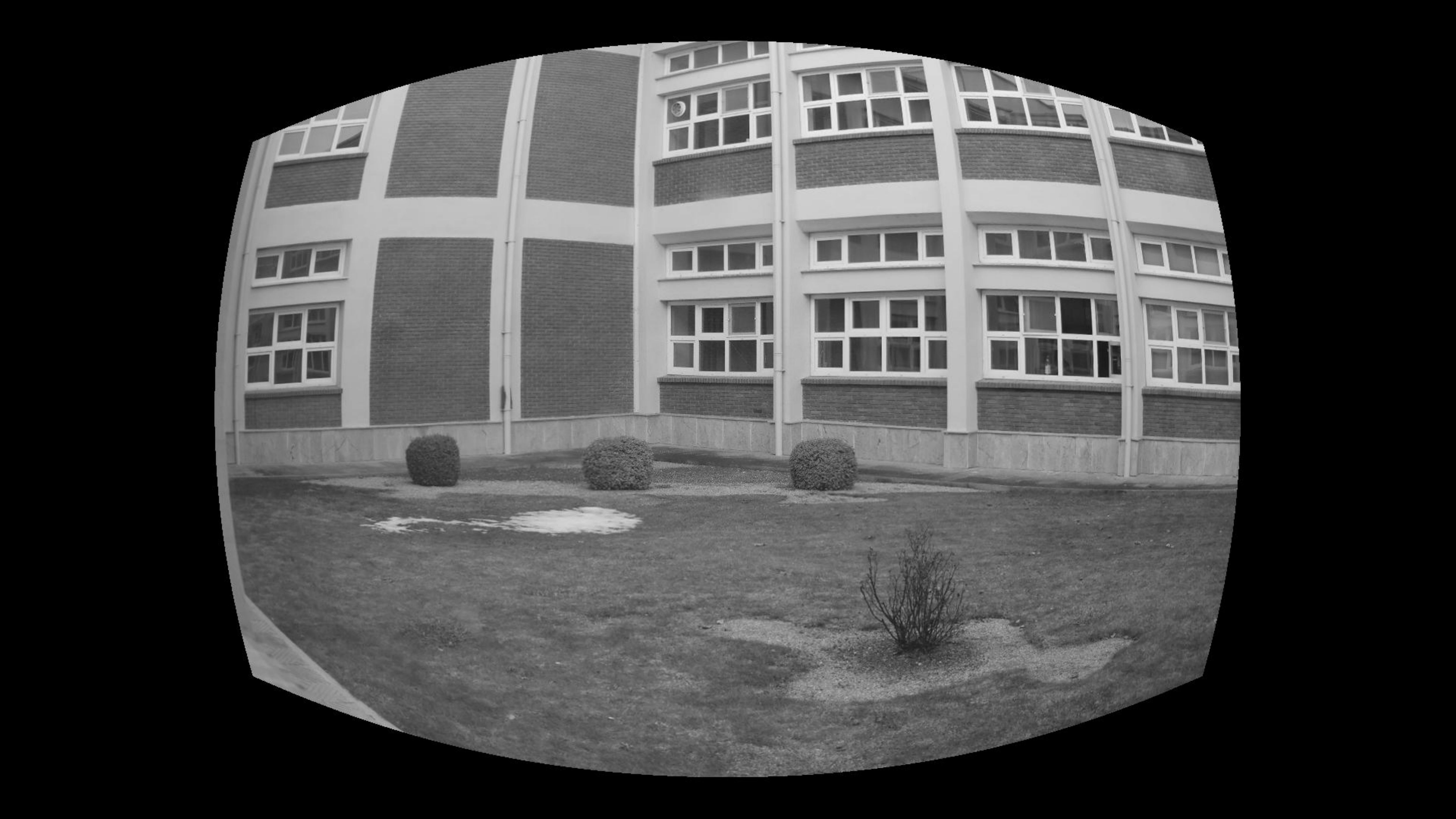}}
		\centerline{(a) $\kappa_{1}^{\prime}=1\times10^{-11}pix^{-2}$}
		\centerline{ $\kappa_{2}^{\prime}=2\times10^{-12}pix^{-4}$}\medskip
	\end{minipage}
	\begin{minipage}[b]{0.48\linewidth}
		\centering
		\centerline{\includegraphics[width=6cm]{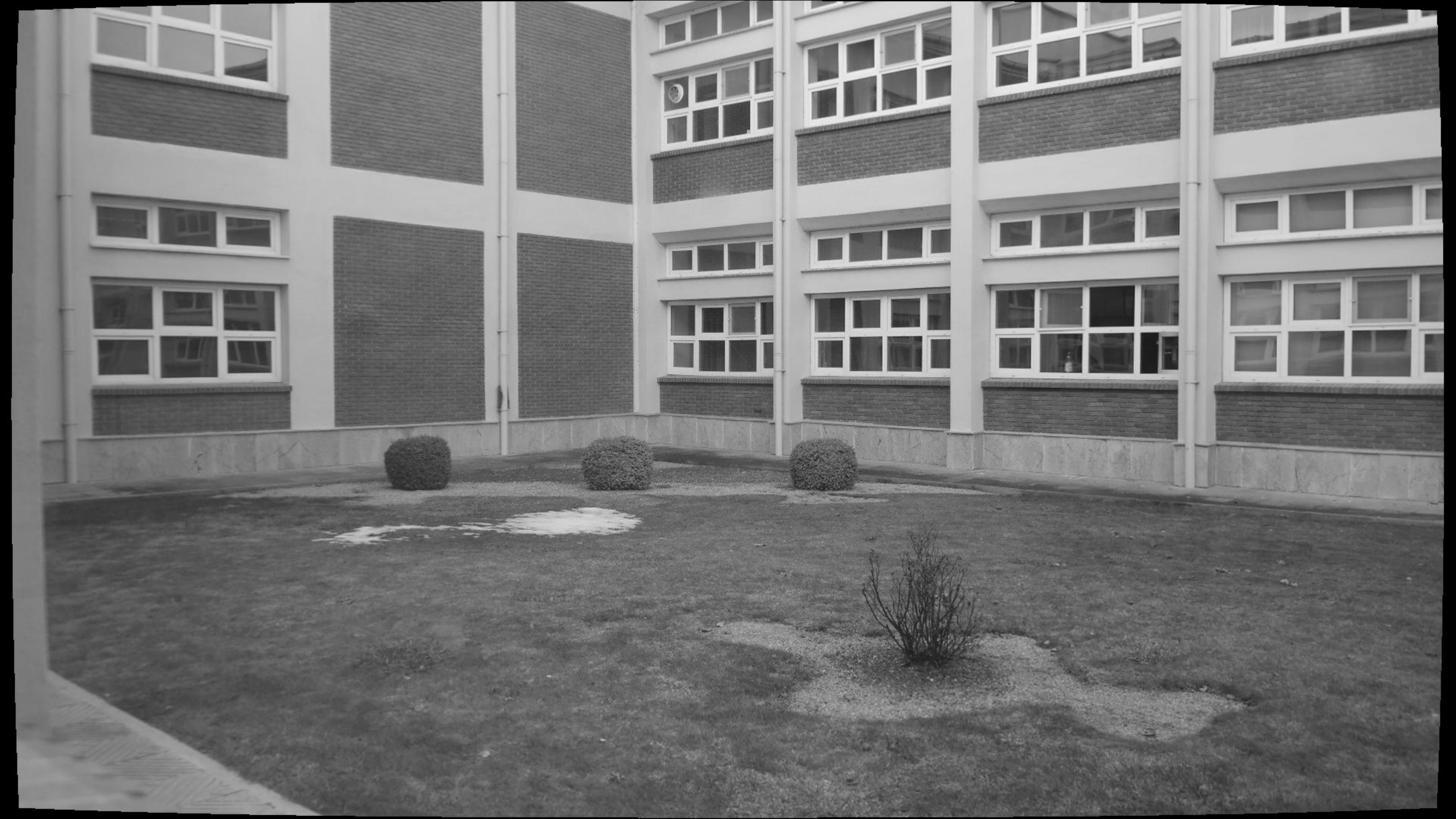}}
		\centerline{(b) $\kappa_{1}^{\prime}=1\times10^{-13}pix^{-2}$}
		\centerline{ $\kappa_{2}^{\prime}=2\times10^{-14}pix^{-4}$}\medskip
	\end{minipage}
	\caption{Examples of artificial distortions applied to an image with $1920\times1080$ resolution.
		The distortion center is assumed to be the center of the image and the ratio of horizontal and vertical focal lengths is unity.}
	\label{fig:dist}
\end{figure}

The composition of scenes or the actual lens distortion model for the test images are not critical, as the images will not be used for calibration.
To avoid any bias, we used the means of results produced from 10 uncompressed photographs taken in indoor and outdoor environments.
We have discussed in Section~\ref{sec:rectification} that forward distortion parameters are used to apply conventional distortion rectification.
In a similar manner, the inverse distortion parameters can be used to apply distortion.
The images are distorted using sets of inverse distortion parameters diverse enough to cover both subtle and substantial distortions.
$\kappa_{1}^{\prime}$ is given values between $1\times10^{-11}pix^{-2}$ and $1\times10^{-13}pix^{-2}$, $\kappa_{2}^{\prime}$ is set to be one fifth of $\kappa_{1}^{\prime}$.
See Figure~\ref{fig:dist} for the minimum and maximum effects of the applied distortion.
Since the signs of parameters do not have a significant effect on the performance of the methods, we omitted the negative range of parameters.

\begin{figure}
	\centering
	\includegraphics[width=\linewidth]{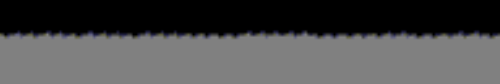}
	\caption{Upper border of an image rectified with the proposed method.
		Triangular patterns appear at the borders of the image due to the interpolation method.
		For the methods that use bilinear interpolation, edges of the images appear as straight lines.}
	\label{fig:edges}
\end{figure}

After distorting the images, the images are rectified using different methods and differences of their results from the original image are quantified using RMSE and PSNR.
The existing methods use bilinear interpolation after approximating the distortion model.
For fair comparison between the methods, we also used bilinear interpolation after triangulation, although more complex interpolation methods such as bicubic are available.
The resulting image is aligned with the input image for all methods.
The interpolation by triangulation method had 2-3 pixel wide artifacts along the borders (see Figure~\ref{fig:edges}), hence these parts are cropped.

\begin{figure}
	\begin{center}
		\includegraphics[width=0.8\linewidth]{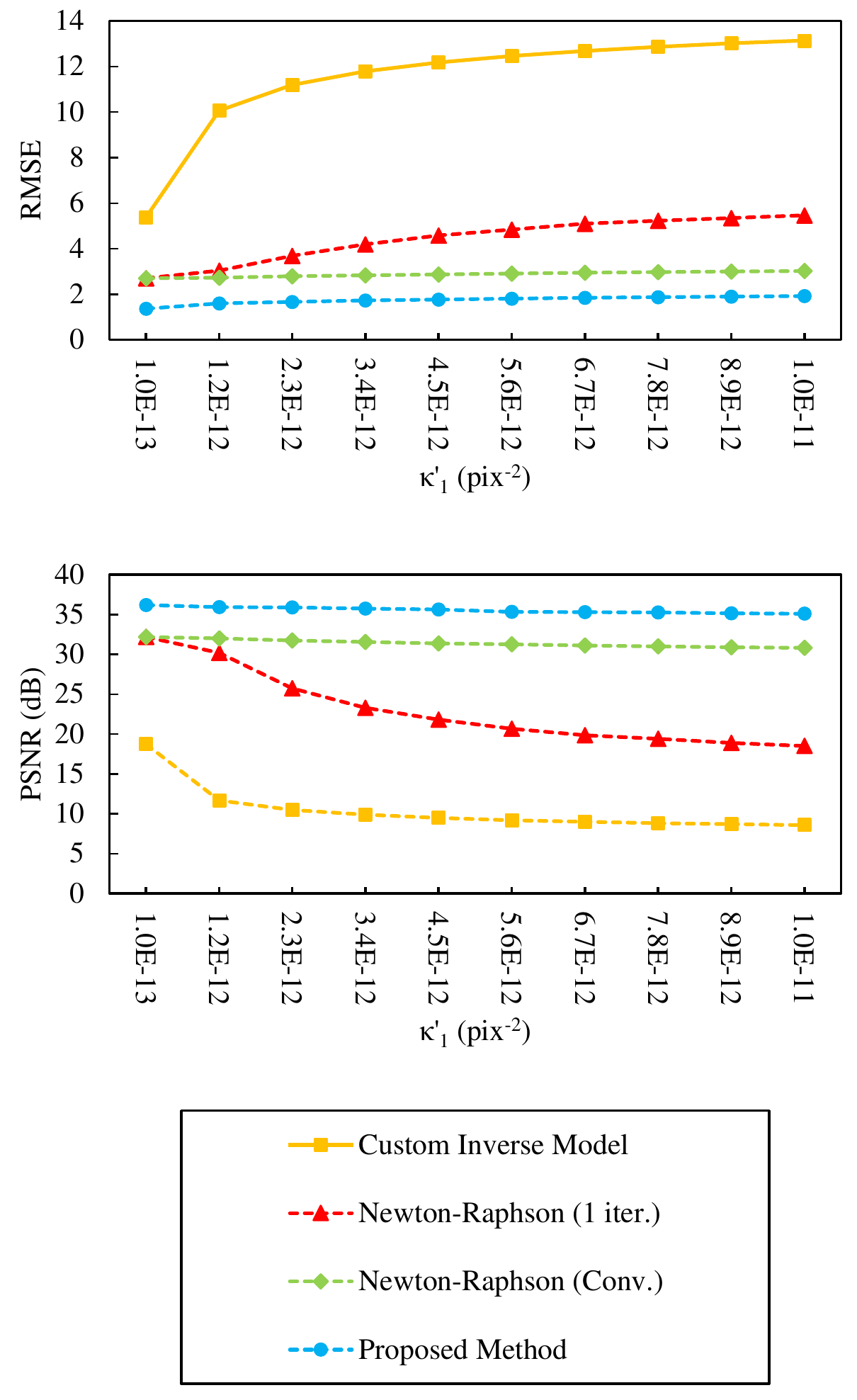}
		\caption{Mean results with a range of inverse distortion coefficients.
			For all steps, $\kappa_{2}^{\prime}$ is set to be $\kappa_{1}^{\prime}/5$.*}
	\end{center}
	*\small Note that differently from the original publication, the horizontal axes are in increasing order.
	In addition, the experiments are repeated and the custom inverse model results are corrected.
	\label{fig:results}
\end{figure}

RMSE and PSNR results for different levels of artificial distortion are given in Figure~\ref{fig:results}.
The proposed method performs stably well across all parameters.
The custom inverse model is approximated by using all normalized pixels on the distorted image in a nonlinear optimization scheme.
Rate of convergence for Newton-Raphson method to a single zero is quadratic.
Starting point is chosen as the undistorted point and the method iteratively converges to its distorted correspondence.
These points are expected to be close to each other, especially where distortion is less pronounced.
Therefore, the convergence is fairly quick, to the point that applying a single iteration of Newton-Raphson may yield acceptable results.
In the experimental results, we chose to show the single iteration case and the case where full convergence is met.
For the distortion coefficients chosen in our experiments, full convergence was met after 5 iterations.
One interesting result of Newton-Raphson method was that it created artifacts on the image along the void patterns shown in Figure~\ref{fig:blank}.
The sampling points are farther away along the patterns, hence the interpolation may have failed to approximate the nonlinearity.
This is not the weakness of Newton-Raphson, as given enough iterations, it should be able to approximate the inverse of the distortion function.
The difference in results between the proposed method and Newton-Raphson originates from the interpolation performance where data points are sparse.

The rectification map is a look up table that lists contributing pixel coordinates from the distorted image for each pixel in the rectified image, along with their respective weights.
This map is used to rectify multiple images with the same distortion characteristics in an efficient manner.
The average running times of different methods for building a rectification map are presented in Table~\ref{tab:times}.
The proposed method gives improved results in a comparable amount of time.
The running time of the proposed method can be improved by processing the rectified points in smaller partitions.
Doing the triangulation operation in small batches will be faster than triangulating the entire image.
Obviously, this may change the final result of the triangulation.
The imposed difference may either be insignificant, or negative.
After the rectification map is built for a distortion model, rectification of images with each method will take the same amount of processing power and memory.

\begin{table}
	\centering
	\caption{Average running times for building the distortion rectification map using a \mbox{$1920\times1080$} image.}
	\tabulinesep=1.2mm
	\begin{tabu}{l c}
		\tabucline[2pt]{-}
		\textbf{Method} & \textbf{Average Running Times (s)} \\
		\hline
		\textbf{Custom Inverse Model} & 76.7 \\
		\textbf{Newton-Raphson (Single Iteration)} & 26.8 \\
		\textbf{Newton-Raphson (Complete Convergence)} & 32.3 \\
		\textbf{Proposed} & 33.4 \\ 
		\tabucline[2pt]{-}
	\end{tabu}
	\label{tab:times}
\end{table}
\vspace{-5mm}

\section{Conclusion}
\label{sec:conclusion}
In this paper, a method of image rectification using directly the inverse distortion model is introduced.
Instead of approximating the forward distortion model, the pixels of the distorted image are mapped to the rectified image.
The resulting point set is triangulated using Delaunay's algorithm.
The pixel values of the rectified image are interpolated according to this triangulation.

The method can be applied to any distortion model, as it only uses the results of the mapping function.
Linear interpolation is chosen as it is commonly used in conventional distortion rectification.
However, interpolation accuracy can be improved by using a nonlinear approach.
Delaunay triangulation is used as it is a well-studied and optimized algorithm.
Memory and processing time tradeoffs can be explored using different implementations.
Furthermore, the effect of using different triangulation algorithms may be significant.

\bibliographystyle{ieeetr}
\bibliography{refs}

\begin{thebibliography}{10}

\bibitem{hartley2003}
R.~Hartley and A.~Zisserman, {\em Multiple View Geometry in Computer Vision}.
\newblock Cambridge University Press, 2003.

\bibitem{brown1966}
D.~C. Brown, ``Decentering distortion of lenses,'' {\em Photometric Eng.},
  vol.~32, no.~3, pp.~444--462, 1966.

\bibitem{fitzgibbon2001}
A.~W. Fitzgibbon, ``Simultaneous linear estimation of multiple view geometry
  and lens distortion,'' in {\em Proc. IEEE Conf. Comput. Vision and Pattern
  Recognition (CVPR)}, 2001.

\bibitem{mallon2004}
J.~Mallon and P.~F. Whelan, ``Precise radial un-distortion of images,'' in {\em
  Proc. IEEE Int. Conf. Pattern Recognition (ICPR)}, 2004.

\bibitem{devernay2001}
F.~Devernay and O.~Faugeras, ``Straight lines have to be straight,'' {\em Mach.
  Vision and Appl. (MVA)}, vol.~13, no.~1, pp.~14--24, 2001.

\bibitem{brauer2001}
C.~Brauer-Burchardt and K.~Voss, ``A new algorithm to correct fish-eye- and
  strong wide-angle-lens-distortion from single images,'' in {\em Proc. Int.
  Conf. Image Process. (ICIP)}, 2001.

\bibitem{wang2009}
A.~Wang, T.~Qiu, and L.~Shao, ``A simple method of radial distortion correction
  with centre of distortion estimation,'' {\em J. Math. Imaging and Vision
  (JMIV)}, vol.~35, no.~3, pp.~165--172, 2009.

\bibitem{bukhari2013}
F.~Bukhari and M.~N. Dailey, ``Automatic radial distortion estimation from a
  single image,'' {\em J. Math. Imaging and Vision (JMIV)}, vol.~45, no.~1,
  pp.~31--45, 2013.

\bibitem{cucchiara2003}
R.~Cucchiara, C.~Grana, A.~Prati, and R.~Vezzani, ``A {H}ough transform-based
  method for radial lens distortion correction,'' in {\em Proc. Int. Conf.
  Image Anal. and Process. (ICIAP)}, 2003.

\bibitem{gonzalez2011}
D.~Gonzalez-Aguilera, J.~Gomez-Lahoz, and P.~Rodr{\'\i}guez-Gonz{\'a}lvez, ``An
  automatic approach for radial lens distortion correction from a single
  image,'' {\em IEEE Sensors J.}, vol.~11, no.~4, pp.~956--965, 2011.

\bibitem{alvarez2009}
L.~Alvarez, L.~G{\'o}mez, and J.~R. Sendra, ``An algebraic approach to lens
  distortion by line rectification,'' {\em J. Math. Imaging and Vision (JMIV)},
  vol.~35, no.~1, pp.~36--50, 2009.

\bibitem{heikkila2000}
J.~Heikkila, ``Geometric camera calibration using circular control points,''
  {\em IEEE Trans. Pattern Anal. Mach. Intell. (TPAMI)}, vol.~22, no.~10,
  pp.~1066--1077, 2000.

\bibitem{heikkila1997}
J.~Heikkila and O.~Silv{\'e}n, ``A four-step camera calibration procedure with
  implicit image correction,'' in {\em Proc. IEEE Conf. Comput. Vision and
  Pattern Recognition (CVPR)}, 1997.

\bibitem{wei1994}
G.-Q. Wei and S.~De~Ma, ``Implicit and explicit camera calibration: Theory and
  experiments,'' {\em IEEE Trans. Pattern Anal. Mach. Intell. (TPAMI)},
  vol.~16, no.~5, pp.~469--480, 1994.

\bibitem{grammatikopoulos2007}
L.~Grammatikopoulos, G.~Karras, and E.~Petsa, ``An automatic approach for
  camera calibration from vanishing points,'' {\em ISPRS J. Photogrammetry and
  Remote Sensing}, vol.~62, no.~1, pp.~64--76, 2007.

\bibitem{ahmed2005}
M.~Ahmed and A.~Farag, ``Nonmetric calibration of camera lens distortion:
  Differential methods and robust estimation,'' {\em IEEE Trans. Image Process.
  (TIP)}, vol.~14, no.~8, pp.~1215--1230, 2005.

\bibitem{thormahlen2003}
T.~Thorm{\"a}hlen, H.~Broszio, and I.~Wassermann, ``Robust line-based
  calibration of lens distortion from a single view,'' in {\em Proc. Mirage
  Conf. Comput. Vision / Comput. Graph. Collaboration for Model-based Imaging,
  Rendering, Image Anal. and Graphical Special Effects}, 2003.

\bibitem{prescott1997}
B.~Prescott and G.~McLean, ``Line-based correction of radial lens distortion,''
  {\em Graphical Models and Image Process.}, vol.~59, no.~1, pp.~39--47, 1997.

\bibitem{strand2005}
R.~Strand and E.~Hayman, ``Correcting radial distortion by circle fitting.,''
  in {\em Proc. Brit. Mach. Vision Conf. (BMVC)}, 2005.

\bibitem{brauer2002}
C.~Brauer-Burchardt and K.~Voss, ``Automatic correction of weak radial lens
  distortion in single views of urban scenes using vanishing points,'' in {\em
  Proc. Int. Conf. Image Process. (ICIP)}, 2002.

\bibitem{lertrattanapanich2002}
S.~Lertrattanapanich and N.~K. Bose, ``High resolution image formation from low
  resolution frames using {D}elaunay triangulation,'' {\em IEEE Trans. Image
  Process. (TIP)}, vol.~11, no.~12, pp.~1427--1441, 2002.

\bibitem{lee1980}
D.-T. Lee and B.~J. Schachter, ``Two algorithms for constructing a {D}elaunay
  triangulation,'' {\em Int. J. Comput. and Inform. Sciences}, vol.~9, no.~3,
  pp.~219--242, 1980.

\bibitem{dyn1990}
N.~Dyn, D.~Levin, and S.~Rippa, ``Data dependent triangulations for piecewise
  linear interpolation,'' {\em IMA J. Numerical Anal.}, vol.~10, no.~1,
  pp.~137--154, 1990.

\bibitem{phong1975}
B.~T. Phong, ``Illumination for computer generated pictures,'' {\em Commun. of
  the ACM}, vol.~18, no.~6, pp.~311--317, 1975.

\end{thebibliography}

\end{document}